\documentclass[conference]{IEEEtran}

\makeatletter
\newcommand*\titleheader[1]{\gdef\@titleheader{#1}}
\AtBeginDocument{%
	\let\st@red@title\@title
	\def\@title{%
		\bgroup\normalfont\large\centering\@titleheader\par\egroup
		\vskip1.5em\st@red@title}
}
\makeatother

\usepackage{tabularx,ragged2e}
\newcolumntype{x}{>{\Centering}X}
\usepackage{hyperref}
\hypersetup{draft}
\usepackage{verbatim}
\usepackage{textcomp}

\usepackage[pdftex]{graphicx}
\usepackage{multicol}
\usepackage{multirow}
\usepackage{caption}
\usepackage{subcaption}

\usepackage{array}
\usepackage{fixltx2e}

\title{Advancements in Image Classification using Convolutional Neural Network}
\titleheader{\textit {This paper has been accepted and presented on 2018 Fourth International Conference on Research in Computational Intelligence and Communication Networks. This is preprint version and original proceeding will be published in IEEE Xplore.}  }
\begin{document}

\author{\IEEEauthorblockN{Farhana Sultana}
\IEEEauthorblockA{Department of Computer Science\\
University of Gour Banga\\
West Bengal, India\\
Email: sfarhana@ieee.org}
\and
\IEEEauthorblockN{Abu Sufian}
\IEEEauthorblockA{Department of Computer Science\\
    University of Gour Banga\\
    West Bengal, India\\
    Email: sufian@ieee.org}
\and
\IEEEauthorblockN{Paramartha Dutta}
\IEEEauthorblockA{Department of CSS\\
Visva-Bharati University\\
West Bengal, India\\
Email: paramartha.dutta@gmail.com}}

\IEEEoverridecommandlockouts
\IEEEpubid{\makebox[\columnwidth]{
		978-1-5386-7638-7/18/\$31.00~\copyright~2018 IEEE \hfill}
	 \hspace{\columnsep}\makebox[\columnwidth]{ }}
	
\maketitle


\renewcommand\IEEEkeywordsname{Keywords}
    
    \begin{abstract}
    Convolutional Neural Network (CNN) is the state-of-the-art for image classification task. Here we have briefly discussed different components of CNN. In this paper, We have explained different CNN architectures for image classification. Through this paper, we have shown advancements in CNN from LeNet-5 to latest SENet model. We have discussed the model description and training details of each model. We have also drawn a comparison among those models. 
       
    \end{abstract}

    \par
    


\section{Introduction}

 Computer vision consists of different problems such as image classification, localization, segmentation and object detection. Among those, image classification can be considered as the fundamental problem and forms the basis for other computer vision problems. Until '90s only traditional machine learning approaches were used to classify image. But the accuracy and scope of the classification task were bounded by several challenges such as hand-crafted feature extraction process etc. In recent years, the deep neural network (DNN), also entitled as deep learning \cite{Lecun15}\cite{Goodfellow16}, finds complex formation in large data sets using the backpropagation \cite{Nielsen89} algorithm. 
 Among DNNs, convolutional neural network has 
 demonstrated excellent achievement in problems of computer vision, especially in image classification. \par

 Convolutional Neural Network (CNN or ConvNet) is a especial type of multi-layer neural network inspired by the mechanism of the optical system of living creatures. Hubel and Wiesel \cite{hubel68} discovered that  animal visual cortex cells detect light in the small receptive field. Motivated by this work, in 1980, Kunihiko Fukushima introduced neocognitron \cite{Fukushima80} which is a 
  multi-layered neural network capable of recognizing visual pattern hierarchically through learning. This network is considered as the theoretical inspiration for CNN. In 1990 LeCun et al. introduced the practical model of CNN \cite{lecun1989} \cite{lecun90} and  developed LeNet-5  \cite{lecun98}. Training by backpropagation \cite{LeCun1988} algorithm helped LeNet-5 recognizing visual patterns from raw pixels directly without using any separate feature engineering mechanism. Also fewer connections and parameters of CNN than conventional feedforward neural networks with similar network size, made model training easier. But at that time in spite of several advantages, the performance of CNN in intricate problems such as classification of high-resolution image, was limited by the lack of large training data, lack of better regularization method and inadequate computing power.\par  
 
 Nowadays we have larger datasets with millions of high resolution labelled data of thousands category like ImageNet \cite{deng09}, LabelMe \cite{russell08} etc. With the advent of powerful GPU machine and better regularization method, CNN delivers outstanding performance on image classification tasks. In 2012 a large deep convolution neural network, called AlexNet \cite{alex12}, designed by Krizhevsky et al. showed excellent performance on the ImageNet Large Scale Visual Recognition Challenge (ILSVRC) \cite{Russakovsky2015}. The success of AlexNet has become the inspiration of different CNN model such as ZFNet \cite{zeiler14}, VGGNet \cite{simonyan14}, GoogleNet \cite{szegedy15}, ResNet \cite{he16}, DenseNet \cite{Huang16}, CapsNet \cite{capsnet17}, SENet \cite{Hu17} etc in the following years. 
 
 In this study, we have tried to give a review of the advancements of the CNN in the area of image classification. We have given a general view of CNN architectures in section II. Section III describes architecture and training details of different models of CNN. In Section IV we have drawn a comparison between various CNN models. Finally, we have concluded our paper in Section V.  


\section{Convolutional Neural Network}
\begin{figure}[htb]
    \centering
    \includegraphics[scale=0.4 ]{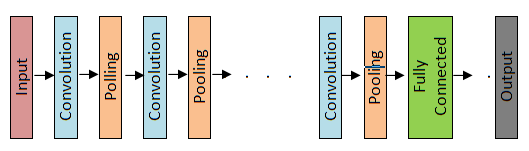}
    \caption{Building block of a typical CNN}
    \label{fcnn}
\end{figure}

A typical CNN is composed of single or multiple blocks of convolution and sub-sampling layers, after that one or more fully connected layers and an output layer as shown in figure \ref{fcnn}.

    \subsection{Convolutional Layer}
\begin{figure}[htb]
    \centering
    \includegraphics[scale=0.3]{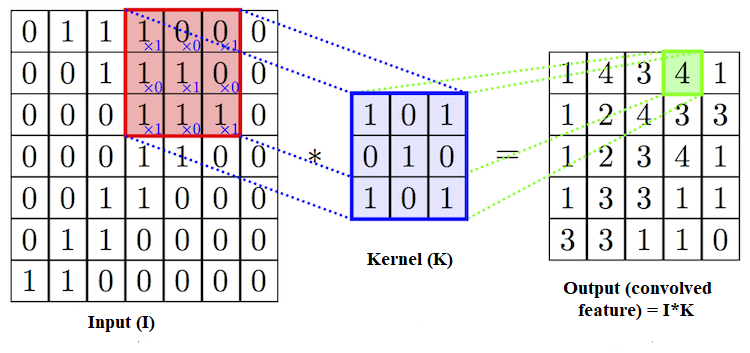}
    \caption{Convolutional Layer}
    \label{fconv_layer}
    \end{figure}

    The convolutional layer (conv layer) is the central part of a CNN. Images are generally stationary in nature. That means the formation of one part of the image is same as any other part. So, a feature learnt in one region can match similar pattern in another region. In a large image, we take a small section and pass it through all the points in the large image (Input). While passing at any point we convolve them into a single position (Output). Each small section of the image that passes over the large image is called filter (Kernel). The filters are later configured based on the back propagation technique. Figure \ref{fconv_layer} shows typical convolutional operation.
     
    \subsection{Sub-sampling or Pooling Layer}
\begin{figure}[htb]
         \centering
         \includegraphics[scale=0.2 ]{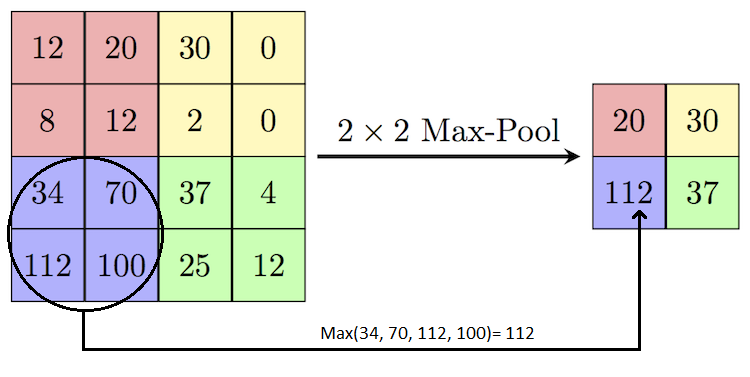}
         \caption{Max Pooling operation}
         \label{fmax_pool}
        \end{figure}

    Pooling simply means down sampling of an image. It takes small region of the convolutional output as input and sub-samples it to produce a single output. Different pooling techniques are there such as max pooling, mean pooling, average pooling etc. Max pooling takes largest of the pixel values of a region as shown in figure \ref{fmax_pool}. Pooling reduces the number of parameter to be computed but makes the network invariant to translations in shape, size and scale.     
    
    \subsection{Fully-connected Layer (FC Layer)}
\begin{figure}[htb]
            \centering
            \includegraphics[scale=0.3 ]{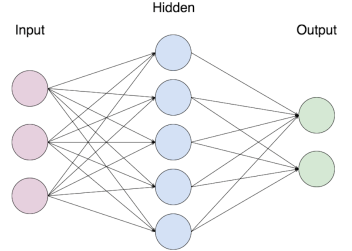}
            \caption{Fully-connected layer}
            \label{ffclayer}
        \end{figure}

    Last section of CNN are basically fully connected layers as depicted in figure \ref{ffclayer}. This layer takes input from all neurons in the previous layer and performs operation with individual neuron in the current layer to generate output. 
    
\section{Different Models of CNN for Image Classification}

\subsection{LeNet-5(1998):}
\begin{figure}[htb]
    \includegraphics[width=\linewidth]{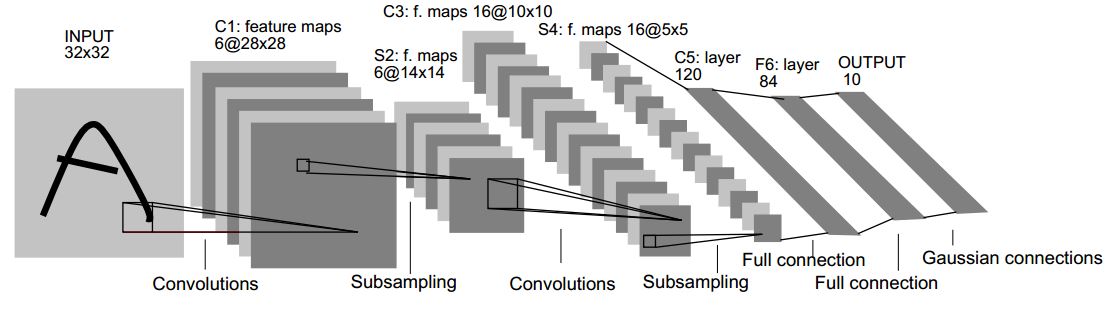}
    \caption{Architecture of LeNet-5 \cite{lecun98}}
    \label{fLeNet5}
\end{figure}

In 1998 LeCun et al. introduced the CNN to classify handwritten digit. Their CNN model, called LeNet-5 \cite{lecun98} as shown in figure \ref{fLeNet5}, has 7 weighted (trainable) layers. Among them, three (C1, C3, C5) convolutional layers, two (S2, S4) average pooling layers, one (F6) fully connected layer and one output layer. 
 $Sigmoid$ function was used to include non-linearity before a pooling operation. The output layer used Euclidean Radial Basis Function units (RBF) \cite{buhmann2000} to classify 10 digits. 

In table \ref{tLeNet5} we have shown different layers, size of the filter used in each convolution layer, output feature map size and the total number of parameters required per layer of LeNet-5.  

\begin{table}[htb]
    \centering
    \caption{Architecture of LeNet-5}
    \begin{tabularx}{\linewidth}{|p{2.4cm}|p{1.1cm}|x|p{1.4cm}|x|}\hline
    \textbf{Layer} &\textbf{filter size/stride} & \textbf{\# filter} & \textbf{output size} & \textbf{\#Para- meters} \\
    \hline
    Convolution(C1) & $5\times5$/1 & 6 & $28\times28\times6$ & 156\\
    \hline
    Sub-sampling(S2) & $2\times2$/2 & & $14\times14\times6$ & 12\\
    \hline
    Convolution(C3) & $5\times5$/1 & 16 & $10\times10\times16$ &  1516\\
    \hline
    Sub-sampling(S4) & $2\times2$/2 & & $5\times5\times16$ & 32\\
    \hline
    Convolution(C5) & $5\times5$ & 120 & $1\times1\times120$ & 48120\\
    \hline
    Fully Connected(F6) & $2\times2$ & & $14\times14\times6$ & 10164\\
    \hline
    OUTPUT & & & & 84\\
    \hline    
    \end{tabularx}
    \label{tLeNet5}
\end{table}
\par

    \subsubsection{Dataset used}
    To train and test LeNet-5, LeCun et al. used the MNIST \cite{mnist10} database of handwritten digits. The database contains 60k training and 10k test data. The input image size of this model is basically $32\times32$ pixels which is larger than the largest character ($20\times20$ pixels) in the database as center part of the receptive field is rich in features. Input images are size normalized and centred in a $28\times28$ field. They have used data augmentation like horizontal translation, vertical translation, scaling, squeezing and horizontal shearing.

    \subsubsection{Training Details}
    The authors trained several versions of LeNet-5 using stochastic gradient descent (SGD) \cite{Leon10} approach with 20 iterations for entire training data per session with a decreased rate of global learning rate and a momentum of 0.02. In 1990's LeNet-5 was sufficiently good. LeNet-5 and LeNet-5 (with distortion)  achieved test error rate of 0.95\% and 0.8\% respectively on MNIST data set.
    
     But as the amount of data, resolution of an image and the number of classes of a classification problem got increased with time, we needed deeper convolutional network and powerful GPU machine to train the model.\par

\subsection{AlexNet-2012:}
   
\begin{figure}[htb]
    \includegraphics[width=\linewidth]{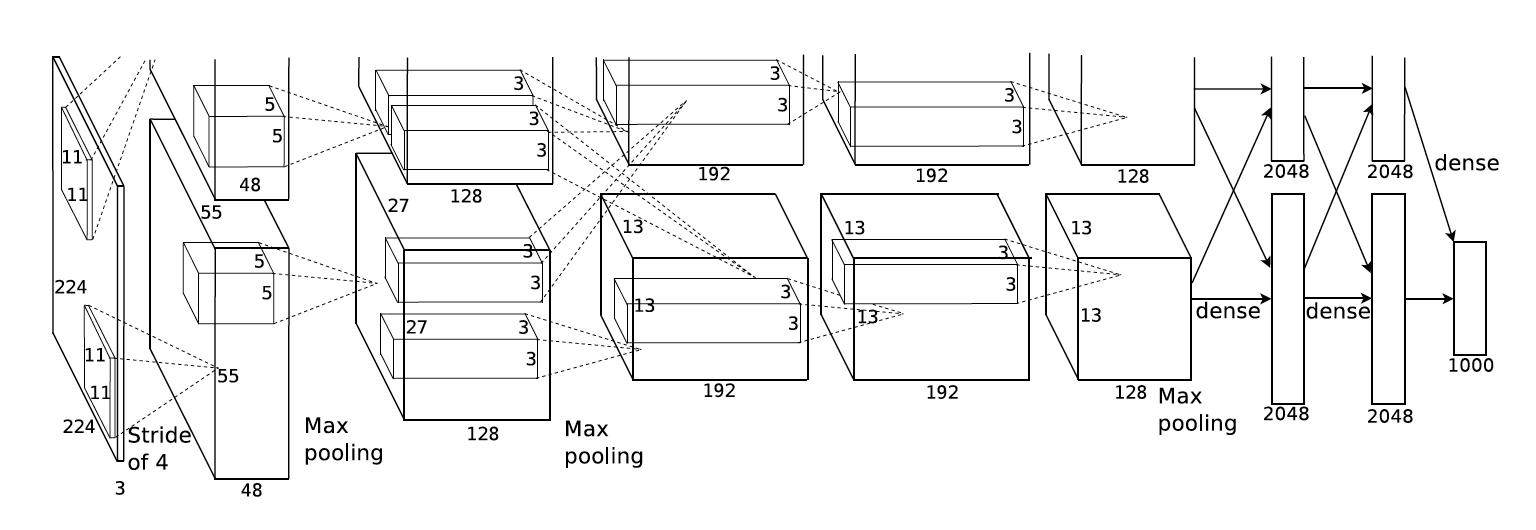}
    \caption{Architecture of AlexNet \cite{alex12} }
    \label{falexnet}
    \end{figure}

In 2012 Krizhevky et al. designed a large deep CNN, called AlexNet \cite{alex12}, to classify ImageNet \cite{deng09} data. The architecture of AlexNet is same as LeNet-5 but much bigger. It is made up of 8 trainable layers. Among them, 5 convolutional layers (conv layer) and 3 fully connected layers are there. Using rectified linear unit (ReLU) \cite{nair10} non-linearity after convolutional and FC layers helped their model to be trained faster than similar networks with $tanh$ units. They have used local response normalization (LRN), called  "brightness normalization",  after the first and second convolutional layer which aids generalization. They have used max-pooling layer after each LRN layer and fifth convolutional layer. In figure \ref{falexnet} architectural details of AlexNet is shown. In table \ref{talexnet} we have shown different elements of AlexNet.       
         
    \begin{table}[htb]
    \centering
    \caption{Details of different layers of AlexNet}
    \begin{tabularx}{\linewidth}{|x|p{1.05cm}|x|x|p{1.55cm}|p{1cm}|}\hline
        \textbf{Layer} &\textbf{filter size/\newline stride} & \textbf{padding} & \textbf{\# filter} & \textbf{output size} & \textbf{\#Para meters} \\
        \hline
        Conv-1 & $11\times11$/4 & 0 & 96 & $55\times55\times96$ & 34848  \\  
        \hline
        pool-1 & $3\times3$/2 & & & $27\times27\times96$ &  \\
        \hline
        Conv-2 & $5\times5$/1 & 2 & 256 & $27\times27\times256$ & 614400 \\
        \hline
        pool-2 & $3\times3$/2 & & & $13\times13\times256$ &   \\
        \hline
        Conv-3 & $3\times3$/1 & 1 & 384 & $13\times13\times384$ & 981504 \\
        \hline
        Conv-4 & $3\times3$/1 & 1 & 384 & $13\times13\times384$ & 1327104 \\
        \hline
        Conv-5 & $3\times3$/1 & 1 & 256 & $13\times13\times256$ & 884736 \\
        \hline
        pool3 & $3\times3$/2 &  &  & $6\times6\times256$ &   \\
        \hline
        FC6 &  &  & & $1\times1\times4096$  & 37748736 \\
        \hline
        FC7 & &  &  & $1\times1\times4096$ & 16777216 \\ \hline
        FC8 & &  &  & $1\times1\times1000$  & 4096000 \\ \hline
    \end{tabularx}
    \label{talexnet}
    \end{table}

   \subsubsection{Dataset used}
   Krizhevsky et al. designed AlexNet for classification of 1.2 million high-resolution images of 1000 classes for ILSVRC - 2010 and ILSVRC - 2012 \cite{ILSVRC15} . There are around 1.2 million/50K/150K training/validation/testing images. On ILSVRC, competitors submit two kinds of error rates: top-1 and top-5. 
  \subsubsection{Training Details}
  From the variable resolution image of ImageNet, AlexNet used down-sampled and centred $256\times256$ pixels image. To reduce overfitting they have used runtime data augmentation as well as a regularization method called dropout \cite{hinton12}. In data augmentation, they have extracted translated and horizontally reflected 10 random patches of $224\times224$ images and also used principal component analysis (PCA) \cite{Jolliffe2011} for RGB channel shifting of training images. The authors trained AlexNet using stochastic gradient descent (SGD) with batch size of 128, weight decay of 0.0005 and momentum of 0.9. The weight decay works as a regularizer and it reduces training error also. Their initial learning rate was 0.01 reduced manually three times by $1/10$ when value accuracy plateaus. AlexNet was trained on two NVIDIA GTX-580 3 GB GPUs using cross-GPU parallelization for five to six days.
  
  The authors have noticed that removing any middle layer degrades network's performance. So, the result depends on the depth of the network. Also, they have used purely supervised learning approach to simplify their experiment, but they have expected that unsupervised pre-training would help if we can have adequate computational power to remarkably increase the network size without increasing the amount of the corresponding labelled dataset.\par

\subsection{ZFNet}
    In 2014 Zeiler and Fergus presented a CNN called ZFNet \cite{zeiler14}. The Architecture of AlexNet and ZFNet is almost similar except that the authors have reduced 1st layer filter size to $7\times7$ instead of $11\times11$ and used stride 2 convolutional layer in both first and second layers to retain more information in those layers' features. In their paper, the authors tried to explain the reason behind the outstanding performance of large deep CNN. They have used a novel visualization technique which is a deconvolutional network with multiple networks, called deconvnet \cite{zeiler11}, to map activation at higher layers back to the space of input pixel to recognize which pixels of the input layer is accountable for a given activation in the feature map. Basically, deconvnet is a reversely ordered convnet. It accepts feature map as input and applies unpooling using a switch. A switch is basically the position of maxima within a pooling region recorded during convolution. Then they rectify it using ReLU non-linearity and uses the transpose version of filters to rebuild the activity in the layer below which activates the chosen activation.

    \begin{figure}[htb]
        \includegraphics[width=\linewidth]{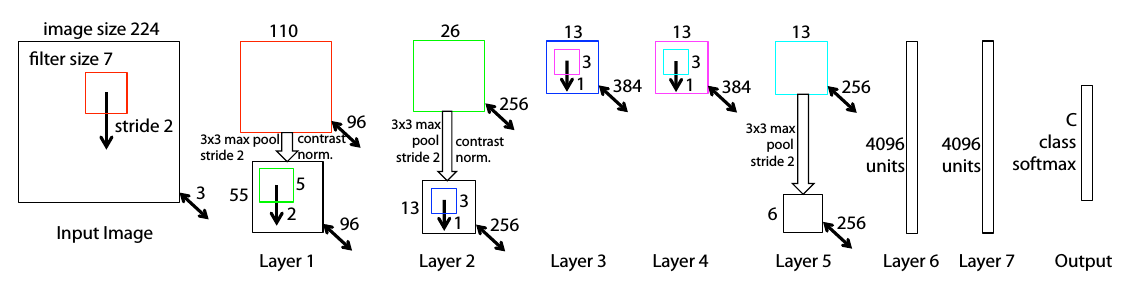}
        \caption{Architecture of ZFNet \cite{zeiler14}}
        \label{fzfnet}
    \end{figure}

           \subsubsection{Training Details}
        ZFNet used the ImageNet dataset of 1.3 million/50k/100k training/validation/testing images. The authors trained their model following \cite{alex12}. The slight difference is that they have substituted the sparse connection of layers 3, 4 and 5 of AlexNet with a dense connection in their model and trained it on single GTX-580 GPU for 12 days with 70 epochs. They have also experimented their model with different depths and different filter sizes on Caltech 101 \cite{caltech101}, Caltech-256 \cite{caltech256} and PASCAL-2012 \cite{pascal2012} data set and shown that their model also generalizes these datasets well.\par
        
        During training their visualization technique discovers different properties of CNN such as the projections from each layer in ascending order shows that the nature of the features are hierarchical in the network. For this reason, firstly, the upper layers need a higher number of epochs than lower layers to converge and secondly, the network output is stable to translation and scaling. They have used a bunch of occlusion experiments to check whether the model is sensitive to local or global information.

\subsection{VGGNet}
\begin{figure}[htb]
	\includegraphics[width=\linewidth]{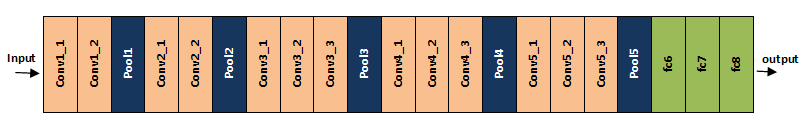}
	\caption{Architecture of VGGNet (configuration D, VGG16)}
	\label{fvggnet}
\end{figure}

Simonyan and Zisserman used deeper configuration of AlexNet \cite{alex12}, and they proposed it as VGGNet \cite{simonyan14}. They have used small filters of size $3\times3$ for all layers and made the network deeper keeping other parameters fixed. They have used total 6 different CNN configurations: A, A-LRN, B, C, D (VGG16) and E (VGG19) with 11, 11, 13, 16, 16, 19 weighted layers respectively. Figure \ref{fvggnet} shows configuration of model D.

The authors have used three $1\times1$ filters in the sixth, ninth and twelfth convolution layer in model C to increase non-linearity. Also, a pack of three $3\times3$ convolution layers (with stride 1) has same effective receptive field as one $7\times7$ convolution layer. So, They have substituted a single $7\times7$ layer with a pack of three $3\times3$ convolution layers and this change increases non-linearity and decreases the number of parameters of the network. 

\subsubsection{Training Details}
The training procedure of VGGNet follows AlexNet except the cropping and scaling sizes of input image for training and testing. 
 Pre-initialization of certain layers and uses of small filters helps their model to converge after 74 epoch in spite of having a large number of parameters and greater depth.
 They have trained configuration VGG A with random initialisation. Then using its first 4 convolution layers and last 3 FC layers as pre-initialised layers, they gradually increased the number of weighted layers up to 19 and trained VGG A-LRN to E. They have randomly cropped image to $224\times224$ from isotropically rescaled training images. They perform horizontal flipping, random RGB colour shifting and scale jittering as data augmentation technique.
The scale jittering in train/test phase, the blending of cropped (multi-crop) and uncropped (dense) test images result in better accuracy.

The authors experienced that a deep network with small filters performs better than a shallower one with larger filters. So the depth of the network is important in visual representation.

\subsection{GoogLeNet}
\begin{figure}[htb]
    	
    	\begin{subfigure}{0.5\textwidth}
    		\centering
    		\includegraphics[width=0.7\linewidth]{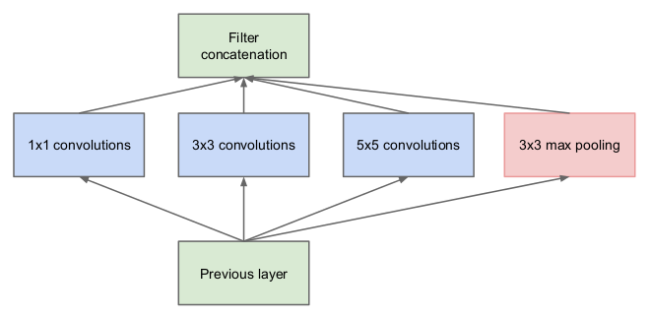} 
    		\caption{}
    		\label{fnaive_inception}
    	\end{subfigure}
    	
    	\begin{subfigure}{0.5\textwidth}
    		\centering
    		\includegraphics[width=0.7\linewidth]{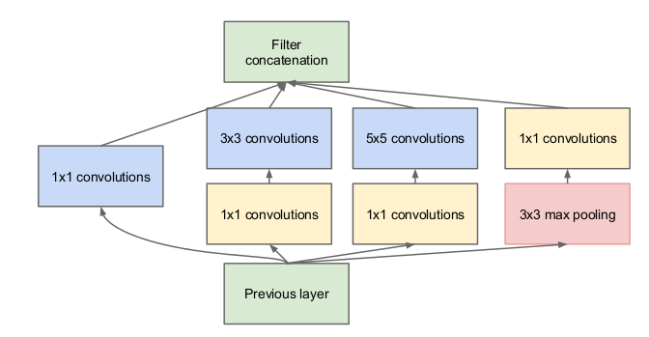}
    		\caption{}
    		\label{finception_module}
    	\end{subfigure}
    	
    	\caption{(a) Naive version Inception Module (b) Dimensionality reduction version \cite{szegedy15}}
    	\label{fnaive_vs_inception}
    \end{figure}

\begin{figure*}[htb]        
          \includegraphics[width=\textwidth]{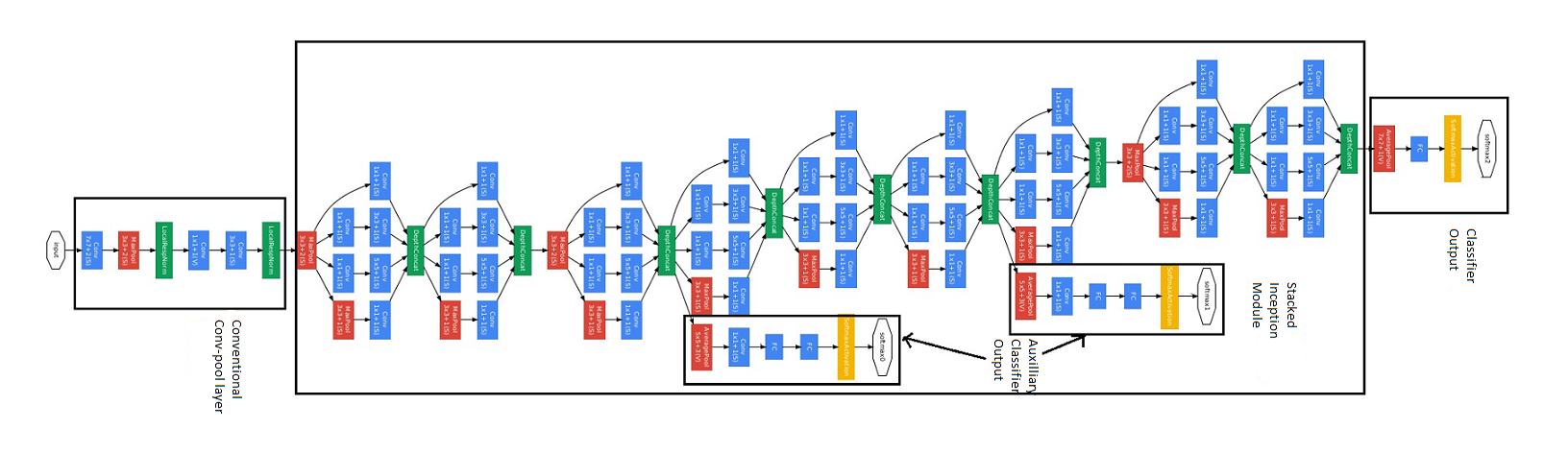}
         \caption{The architecture of GoogLeNet \cite{szegedy15}}
        \label{fgooglenet}
         \end{figure*}

    The architecture of GoogLeNet \cite{szegedy15}, proposed by Szegedy et al., is different from conventional CNN. They have increased the number of units in each layer using parallel filters called inception module \cite{Lin13} of size $1\times1$, $3\times3$ and $5\times5$ in each convolution layer (conv layer). They have also increased the layers to 22. Figure \ref{fgooglenet} shows the 22 layers GoogLeNet. While designing this model, they have considered the computational budget fixed. So that the model can be used in mobile and embedded systems. They have used a series of weighted Gabor filters \cite{serre07} of various size in the inception architecture to handle multiple scales. To make the architecture computationally efficient they have used inception module with dimensionality reduction instead of the naive version of inception module. Figure \ref{fnaive_inception} and figure \ref{finception_module} are showing both inception modules. Despite 22 layers, the number of parameters used in GoogLeNet is 12 times lesser than AlexNet but its accuracy is significantly better. All the convolution, reduction and projection layers use ReLU non-linearity. They have used average pooling layer instead of the fully connected layers. On top of some inception modules, they have used auxiliary classifiers which are basically smaller CNNs, to combat vanishing gradient problem and overfitting.

          \subsubsection{Training Details}
          GoogLeNet, a CPU based implementation, was trained using DistBelief \cite{dean12} distributed machine learning system by using moderate amount of model and data parallelization. They used asynchronous SGD with momentum 0.9 and a constant learning rate schedule. Using different sampling and random ordering of input images, they have trained 7 ensemble GoogLeNet with same initialization. Unlike AlexNet they have used resized image of 4 scales with shorter dimension of 256, 288, 320 and 352 respectively. The total number of crops per image is 4 (scales) $\times 3$ (left, right and centre square/scale) $\times 6$ (4 corner and centre $224\times224$ crop and the square resized to $224\times224$) $\times2$ (mirror image of all six crops)=144.    
          
          The result of inception architecture has proved that moving towards sparser architecture is realistic and competent idea. 
          
\subsection{ResNet}      
    He et al. experienced that a deeper CNN stacked up with more layers suffers from vanishing gradient problem. Though this problem is handled by normalized and intermediate initialization, the deeper model shows worse performance on both train and test errors and it is not caused by overfitting. This indicates that optimization of deeper network is hard. To solve this problem the authors used pre-trained shallower model with additional layers to perform identity mapping. So that the performance of deeper network and the shallower network should be similar. They have proposed deep residual learning framework \cite{he16} as a solution to the degradation problem. They have included residual mapping ($ H(x)=F(x)+x$) instead of desired underlying mapping ($H(x)$) into their network and named their model as ResNet \cite{he16}. 
            
       \begin{figure}[htb]
       	
       	\begin{subfigure}{0.45\linewidth}
       		\centering
       		\includegraphics[width=0.5\textwidth]{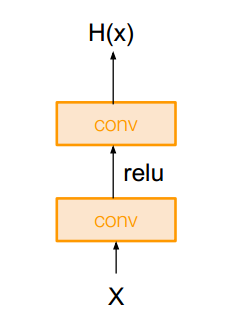} 
       		\caption{}
       		\label{fplain_layer}
       	\end{subfigure}
       	\quad
       	\begin{subfigure}{0.45\linewidth}
       		\centering
       		\includegraphics[width=0.8\textwidth]{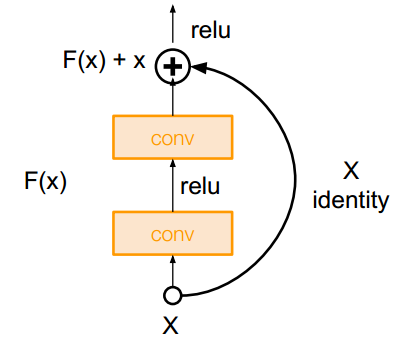}
       		\caption{}
       		\label{fres_block}
       	\end{subfigure}
       	
       	\caption{(a) Plain layer (b) Residual block \cite{he16}}
       	\label{fplain_vs_res}
       \end{figure}
   
    
    ResNet architecture consists of stacked residual blocks of $3\times3$ convolutional layers. They have periodically doubled the number of filters and used a stride of 2. Figure \ref{fplain_layer} and \ref{fres_block} shows a plain layer and residual block.
    As a first layer, they have used a $7\times7$ conv layer. They have not used any fully connected layers at the end. They have used different depth (34, 50, 101 and 152) ResNet in ILSVRC-2014 competition. For the CNN with depth more than 50 they have used 'bottleneck' layer for dimensionality reduction and to improve efficiency as GoogLeNet. Their bottleneck design consists of $1\times1$, $3\times3$ and $1\times1$ convolution layer. Although the 152 Layer ResNet is 8 times deeper than VGG nets, it has lower complexity than VGG nets (16/19). 
    
    \subsubsection{Training Details}
     To train ResNet, He et al. used SGD with batch size of 128, weight decay of 0.0001 and momentum of 0.9. They have used a learning rate of 0.1 reduced manually two times at 32k and 48k iterations by $1/10$ when value accuracy plateaus and stopped at 64k iterations. They used weight initialization and Batch Normalization after every conv layer. The did not use dropout regularization method. 
        
     The experiment of ResNet shows the ability to train deeper network without degrading the performance. The authors have also shown that with increased depth the ResNet, it is easier to optimize and it gains accuracy.

\subsection{DenseNet}
\begin{figure}[htb]
    	\centering
    	\includegraphics[scale=0.5]{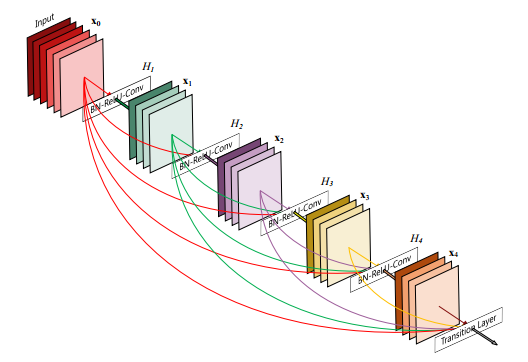}
    	\caption{A typical dense block with 5 layers \cite{Huang16}}
    	\label{fdense_block}
    \end{figure}

    Huang et al. introduced Dense Convolutional Networks (DenseNet) \cite{Huang16}, which includes dense block in conventional CNN. The input of a certain layer in a dense block is the concatenation of the output of all the previous layers as shown in figure \ref{fdense_block}. Here, each layer is reusing the features of all previous layers, strengthening feature propagation and reducing vanishing gradient problem. Also uses of small number of filters reduced the number of parameters as well. 

    \begin{figure}[h]
    	\centering
        \includegraphics[scale=0.5]{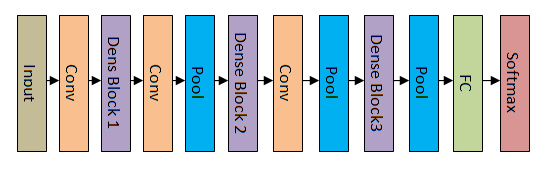}
        \caption{A DenseNet with 3 Dense Block}
        \label{fdensenet}
    \end{figure}
    
    Figure \ref{fdensenet} shows a DenseNet with three dense blocks.  In a dense block, the non-linear transformation functions are a composite function of batch normalization, ReLU and $3\times3$ convolution operation. They have also used the $1\times1$ bottleneck layer to reduce dimensionality. 
    
    \subsubsection{Training Details}     
     Huang et al. trained DenseNet on CIFAR \cite{CIFAR}, SVHN \cite{SVHN11} and ImageNet dataset using SGD with batch size 64 on both CIFAR and SVHN dataset, and with batch size 256 on ImageNet dataset. Initial learning rate was 0.1 and is decreased two times by $1/10$. They have used weight decay of 0.0001, Nesterov momentum \cite{sutskever13} of 0.9 and dropout of 0.2. 
            
     On C10 \cite{CIFAR10}, C100 \cite{CIFAR100}, SVHN dataset DenseNet, DenseNet-BC outperforms the error rates of previous CNN architectures. A DenseNet, doubly deeper than ResNet, gives similar accuracy on ImageNet datasets with very less (factor of 2) number of parameters. The authors experienced that DenseNet can be scaled to hundreds layers without optimization difficulty. It also gives consistent improvement if number of parameters increases without degrading performance and overfitting. Also, it requires comparatively fewer parameters and less computational power for better performance.
     
    \subsection{CapsNet}
    Conventional CNNs, described above, suffer from two problems. Firstly, Sub-sampling loses the spatial information between higher-level features. Secondly, it faces difficulty in generalizing to novel view points. It can deal with translation  but can not detect different dimension of affine transformation. In 2017, Geoffrey E. Hinton proposed CapsNet \cite{capsnet17} to handle these problems. CapsNet has components called capsule. A capsule is a group of neurons. So a layer of CapsNet is basically composed with nested neurons. Unlike a typical neural network, a capsule is squashed as a whole vector rather than individual output unit squashing. So scalar output feature detector of CNN is replaced by vector output capsules. Also max-pooling is replaced by "dynamic routing by agreement" which makes each capsule in each layer to go to the next most relevant capsules at the time of forward propagation. 
    
\begin{figure}[htb]
    	\centering
   	\includegraphics[scale=0.2]{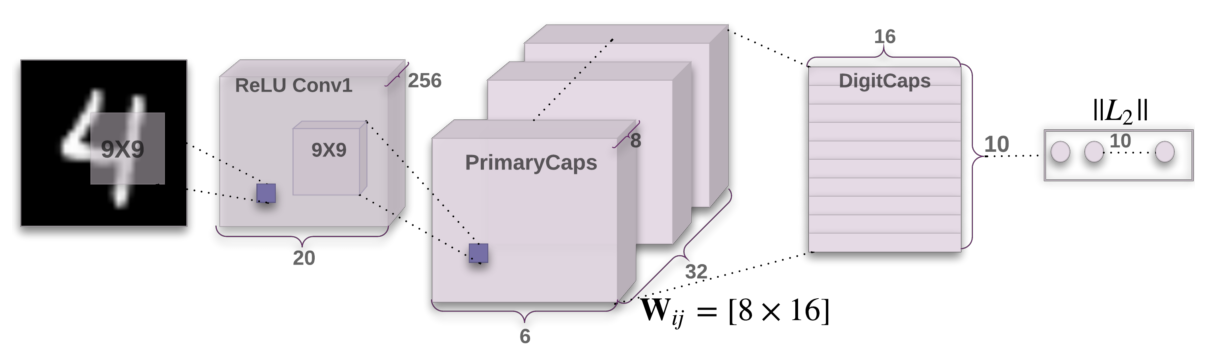}
   	\caption{A 3 layer CapsNet, used for handwritten digit recognition \cite{capsnet17}}
   	\label{fcapsnet}
   \end{figure}

   Architecture of a simple CapsNet is shown in figure \ref{fcapsnet}. 
   
   The CapsNet, proposed by Sabour et al, is composed with three layers - two conv layers and one FC layer. First conv layer consist of 256 convolutional unit (CU) with $9\times9$ kernels of stride 1 and uses ReLU as activation function. This layer detects local features and then sends it to the primary capsules of second layer as input. Each primary capsule contains 8 CU with $9\times9$ kernel of stride of 2. In total primary capsule layer has $32\times6\times6$ 8D capsules. The final layer (DigitCaps) has one 16D capsule per digit class. The authors have used routing between primary layer and DigitCaps layer. As the first convolutional layer is a 1D layer, no routing is used between this layer and primary capsule layer. 
   
   \begin{table*}[t]
   	\centering
   	\caption{Comparative performance of different CNN configurations. 
   		 The + indicates- DenseNet with Bottleneck layer and compression (10 crop testing result). } 
   	\begin{tabularx}{\textwidth}{|p{1cm}|p{1cm}|p{0.5cm}|p{6.8cm}|x|p{0.9cm}|p{0.9cm}|p{0.9cm}|}
   		\hline
   		
   		\textbf{Name of The CNN}&\textbf{Dataset} &\textbf{Year} &\textbf{Type of CNN} & \textbf{\#trained layer} &  \textbf{Top-1(val)} &\textbf{Top-5(val)} &  \textbf{Top-5(test)} \\
   		\hline 
   		\multirow{4}{*}{AlexNet} & \multirow{4}{*}{ImageNet} & \multirow{4}{*}{2012} & 1 CNN & 8 &  40.7\% & 18.2\% &  \\
   		\cline{4-8}  	
   		&  &  & 5 CNN & - &  38.1\% & 16.4\% &16.4\%  \\
   		\cline{4-8}
   		&  &  & 1 CNN & - &  39.0\% & 16.6\% &- \\
   		\cline{4-8}
   		&  &  & 7 CNN & - & 36.7\% & 15.4\% &\textbf{15.3\%}   \\
   		\hline
   		
   		\multirow{4}{*}{ZFNet}& \multirow{4}{*}{ImageNet} & \multirow{4}{*}{2013} & 1 CNN & 8 &  38.4 \% & 16.5\% &   \\
   		\cline{4-8}
   		&  &  & 5 CNN - (a) & - & 36.7 \% & 15.3\% &  15.3\% \\
   		\cline{4-8}  	
   		&  &  & 1 CNN with layers 3, 4, 5: 512, 1024, 512 maps-(b) & - &  37.5 \% & 16.0\% &  16.1\% \\
   		\cline{4-8}  	
   		&  &  & 6 CNN, combination of (a) \& (b) & - &  36.0 \% & 14.7\% &  \textbf{14.8}\% \\     
   		\hline
   		\multirow{4}{*}{VGGNet} & \multirow{4}{*}{ImageNet} & \multirow{4}{*}{2014} & ensemble of 7 ConvNets (3-D,2-C \& 2-E) & - & 24.7\% & 7.5\% &  \textbf{7.3\%} \\  
   		\cline{4-8}   
   		&  &  &ConvNet- D( multi-crop \& dense) & 16 &  24.4 \% & 7.2\% &-  \\
   		\cline{4-8}
   		&  &  & ConvNet-E (Multi-crop \& dense ) & 19 &  24.4 \% & 7.1\% &  - \\
   		\cline{4-8}  
   		&  &  & ConvNet-E (Multi-crop \& dense ) & 19 &  24.4 \% & 7.1\% & 7.0\% \\
   		\cline{4-8}	
   		&  &  & Ensemble of multi-scale ConvNets D \& E (multi-crop \& dense)   & - &  23.7\%  & 6.8\% &  6.8\% \\  	
   		\hline  
   		
   		\multirow{4}{*}{GoogLeNet} & \multirow{4}{*}{ImageNet} & \multirow{4}{*}{2014} & 1 CNN with 1 crop & 22 & - & - & 10.07\%   \\  
   		\cline{4-8}		      				
   		&  &  & 1 CNN with 10 crops  & - &  -  & - &  9.15\% \\
   		\cline{4-8}		      				
   		&  &  & 1 CNN with 144 crops  & - &  -  & - &  7.89\% \\
   		\cline{4-8}		      				
   		&  &  & 7 CNN with 1 crop   & - &  -  & - &  8.09\% \\ 
   		\cline{4-8}		      				
   		&  &  & 1 CNN with 10 crops  & - &  -  & - &  7.62\% \\ 
   		\cline{4-8}		      				
   		&  &  & 1 CNN with 144 crops  & - &  -  & - &  \textbf{6.67\%} \\	
   		\hline	
   		\multirow{4}{*}{ResNet} & \multirow{4}{*}{ImageNet} & \multirow{4}{*}{2015} & plain layer & 18 &  27.94\% & - &    \\  
   		\cline{4-8}	
   		&  &  & ResNet-18  & 18 &  27.88\%  & - &   \\ 
   		\cline{4-8}
   		&  &  & Plain layer  & 34 & 28.54\%   & 10.02 &  \\ 
   		\cline{4-8}		    				
   		&  &  & ResNet-34 (zero-padding shortcuts), 10 crop testing -(a)  & 34 &   25.03\%  & 7.76 &  \\ 
   		\cline{4-8}	
   		&  &  & ResNet-34 (projection shortcuts to increase dimension, others are identity shortcuts ), 10 crop testing-(b) & 34 &  24.52\%  & 7.46\% &   \\ 
   		\cline{4-8}	
   		&  &  & ResNet-34 (all  shortcuts are projection), 10 crop testing-(c)  & 34 &  24.52\%  & 7.46\% &   \\ 
   		\cline{4-8}
   		&  &  & ResNet-50 (with bottleneck layer), 10 crop testing  & 50 &  22.85\% & 6.71\% &   \\ 
   		\cline{4-8}
   		&  &  & ResNet-101 (with bottleneck layer), 10 crop testing  & 101 &  21.75\% & 6.05\% &   \\ 
   		\cline{4-8}
   		&  &  & ResNet-152 (with bottleneck layer), 10 crop testing  & 152 &  21.43\% & 5.71\% &   \\ 
   		\cline{4-8}
   		&  &  & 1 ResNet-34 (b)   & 34 &  21.84\% & 5.71\% &   \\ 
   		\cline{4-8}
   		&  &  & 1 ResNet-34 (c)   & 34 &  21.53\% & 5.60\% &   \\ 
   		\cline{4-8}
   		&  &  & 1 ResNet-50   & 50 &  20.74\% & 5.25\% &   \\ 
   		\cline{4-8}
   		&  &  & 1 ResNet-101   & 101 &  19.87\% & 4.60\% &   \\ 
   		\cline{4-8}
   		&  &  & 1 ResNet-152  & 152 &  19.38\% & 4.49\% &   \\ 
   		\cline{4-8}
   		&  &  & Ensemble of 6 models   & - &   &  & \textbf{3.57\%} \\ 
   		\hline
   		
   		\multirow{4}{*}{DenseNet} & \multirow{4}{*}{ImageNet} & \multirow{4}{*}{2016} & DensNet-121 + & 121 &  23.61\% & 6.66\% &    \\  
   		\cline{4-8}
   		&  &  & DenseNet-169 +  & 169 &  22.80\% & 5.92\% &   \\ 
   		\cline{4-8}
   		&  &  & DenseNet-201 +  & 201 &  22.58\% & 5.54\% &   \\ 
   		\cline{4-8}
   		&  &  & DenseNet-264 +  & 264 &  20.80\% & 5.29\% &   \\ 
   		\hline
   		\multirow{4}{*}{SENet} & \multirow{4}{*}{ImageNet} & \multirow{4}{*}{2017} & SE-ResNet-50 & 50 &23.29\% & 6.62\% &    \\  
   		\cline{4-8}
   	    & &  & SE-ResNext-50   & 50 &  21.10\% & 5.49\% &   \\ 
   			\cline{4-8}
   		& &  & SENet-154 (crop size $320\times320/299\times229$)   &-  &  17.28\% & 3.79\% &   \\ 
   			\cline{4-8}
   		& &  & SENet-154(crop size $320\times320$)   & - & 16.88\% & 3.58\% &   \\ 
   		
   		\hline
   		
   	\end{tabularx}
   	\label{compare_table}
   \end{table*}
   
   \subsubsection{Training details}
   Training of CapsNet is performed on MNIST images. 
   To compare the test accuracy, they have used one standard CNN (baseline)  and two CapsNets with 1 and 3 routing iterations respectively. They have used reconstruction loss as regularization method. Using a 3 layer CapsNet with 3 routing iterations and with added reconstruction the authors get a test error of 0.25\%.
   
   
   Though CapsNet has shown outstanding performance on MNIST, it may not perform well with large scale image dataset like ImageNet. It may also suffer from vanishing gradient problem. 

 \subsection{SENet}
\begin{figure}[htb]
 	\centering
 	\includegraphics[width=\linewidth]{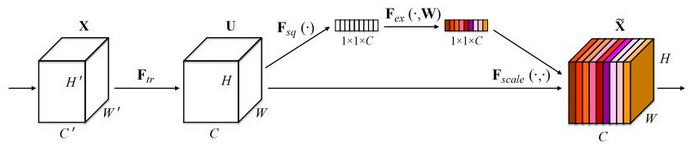}
 	\caption{A Sqeeze and excitation block \cite{Hu17}}
 	\label{fSEnet}
 \end{figure}

 In 2017, Hu et al. have designed "Squeeze-and-Excitation network" (SENet) \cite{Hu17} and have become the winner of ILSVRC-2017. They have reduced the top-5 error rate to 2.25\%. Their main contribution is "Squeeze-and-Excitation" (SE) block as shown in figure \ref{fSEnet}. Here, $F_{tr}$: X$\,\to\,$U is a convolutional operation.
 A squeeze function ($F_{sq}$) performs average pooling on individual 
channel of feature map U and produce $1\times1\times C$ dimensional channel descriptor. An excitation function ($F_{ex}$) is a self-gating mechanism made up of three layers - two fully connected layers and a ReLU non-linearity layer in between. It takes squeezed output as input and produce a per channel modulation weights. By applying the excited output on the feature map U, U is scaled ($F_scale$) to generate final output ($\widetilde{X}$) of SE block.  
 
 This SE block can be stacked together to make SENet which generalise different data set very well. The authors developed different SENets including these blocks into several complex CNN models such as VGGNet \cite{simonyan14}, GoogLeNet \cite{szegedy15}, ResNext (Variant of ResNet) \cite{Xie16}, Inception-ResNet \cite{SzegedyIV16}, MobileNet \cite{Howard17}, ShuffleNet \cite{Zhang17}.  
 
 \subsubsection{Training Details}
 The authors have trained and test their model variants on ImageNet, CIFAR-10 and CIFAR-100. They have trained original CNN models and those models with SE blocks, and compare speed accuracy trade-off. They have shown that their models outperform original models by increasing a little bit training/testing time. 
 

\section{Comparative Result} In table \ref{compare_table}, we have shown comparative performance of different CNN (AlexNet to DenseNet) on ImageNet dataset. Top-1 and top-5 error rate on validation dataset and top-5 error rates on test dataset are also shown.
 

\section{Conclusion}
In this study, we have discussed the advancements of CNN in image classification tasks. We have shown here that although AlexNet, ZFNet and VGGNet followed the architecture of conventional CNN model such as LeNet-5 their networks are larger and deeper. We have experienced that combining inception module and residual blocks with conventional CNN model, GoogLeNet and ResNet gained better accuracy than stacking the same building blocks again and again. DenseNet focused on feature reusing to strengthen the feature propagation. Though CapsNet reached state-of-the-art achievement on MNIST but it is yet to perform as well as previous CNNs performance on high resolution image dataset such as ImageNet. 
 The result of SENet on ImageNet dataset gives us the hope that it may turn out useful for other task which requires strong discriminative features.

\bibliographystyle{IEEEtran}

\bibliography{fsbib}

\begin{thebibliography}{10}
\providecommand{\url}[1]{#1}
\csname url@samestyle\endcsname
\providecommand{\newblock}{\relax}
\providecommand{\bibinfo}[2]{#2}
\providecommand{\BIBentrySTDinterwordspacing}{\spaceskip=0pt\relax}
\providecommand{\BIBentryALTinterwordstretchfactor}{4}
\providecommand{\BIBentryALTinterwordspacing}{\spaceskip=\fontdimen2\font plus
\BIBentryALTinterwordstretchfactor\fontdimen3\font minus
  \fontdimen4\font\relax}
\providecommand{\BIBforeignlanguage}[2]{{%
\expandafter\ifx\csname l@#1\endcsname\relax
\typeout{** WARNING: IEEEtran.bst: No hyphenation pattern has been}%
\typeout{** loaded for the language `#1'. Using the pattern for}%
\typeout{** the default language instead.}%
\else
\language=\csname l@#1\endcsname
\fi
#2}}
\providecommand{\BIBdecl}{\relax}
\BIBdecl

\bibitem{Lecun15}
Y.~Lecun, Y.~Bengio, and G.~Hinton, ``\BIBforeignlanguage{English (US)}{Deep
  learning},'' \emph{\BIBforeignlanguage{English (US)}{Nature}}, vol. 521, no.
  7553, pp. 436--444, 5 2015.

\bibitem{Goodfellow16}
I.~Goodfellow, Y.~Bengio, and A.~Courville, \emph{Deep Learning}.\hskip 1em
  plus 0.5em minus 0.4em\relax MIT Press, 2016,
  \url{http://www.deeplearningbook.org}.

\bibitem{Nielsen89}
R.~Hecht-Nielsen, ``Theory of the backpropagation neural network,'' in
  \emph{International 1989 Joint Conference on Neural Networks}, 1989, pp.
  593--605 vol.1.

\bibitem{hubel68}
D.~H. Hubel and T.~N. Wiesel, ``Receptive fields and functional architecture of
  monkey striate cortex,'' \emph{Journal of Physiology (London)}, vol. 195, pp.
  215--243, 1968.

\bibitem{Fukushima80}
\BIBentryALTinterwordspacing
K.~Fukushima, ``Neocognitron: A self-organizing neural network model for a
  mechanism of pattern recognition unaffected by shift in position,''
  \emph{Biological Cybernetics}, vol.~36, no.~4, pp. 193--202, Apr 1980.
  [Online]. Available: \url{https://doi.org/10.1007/BF00344251}
\BIBentrySTDinterwordspacing

\bibitem{lecun1989}
Y.~LeCun, B.~Boser, J.~S. Denker, D.~Henderson, R.~E. Howard, W.~Hubbard, and
  L.~D. Jackel, ``Backpropagation applied to handwritten zip code
  recognition,'' \emph{Neural Computation}, vol.~1, no.~4, pp. 541--551, Dec
  1989.

\bibitem{lecun90}
\BIBentryALTinterwordspacing
Y.~LeCun, B.~E. Boser, J.~S. Denker, D.~Henderson, R.~E. Howard, W.~E. Hubbard,
  and L.~D. Jackel, ``Handwritten digit recognition with a back-propagation
  network,'' in \emph{Advances in Neural Information Processing Systems 2},
  D.~S. Touretzky, Ed.\hskip 1em plus 0.5em minus 0.4em\relax Morgan-Kaufmann,
  1990, pp. 396--404. [Online]. Available:
  \url{http://papers.nips.cc/paper/293-handwritten-digit-recognition-with-a-back-propagation-network.pdf}
\BIBentrySTDinterwordspacing

\bibitem{lecun98}
Y.~Lecun, L.~Bottou, Y.~Bengio, and P.~Haffner, ``Gradient-based learning
  applied to document recognition,'' \emph{Proceedings of the IEEE}, vol.~86,
  no.~11, pp. 2278--2324, Nov 1998.

\bibitem{LeCun1988}
Y.~L. Cun, ``A theoretical framework for back-propagation,'' 1988.

\bibitem{deng09}
J.~Deng, W.~Dong, R.~Socher, L.-J. Li, K.~Li, and L.~Fei-Fei, ``{ImageNet: A
  Large-Scale Hierarchical Image Database},'' in \emph{CVPR09}, 2009.

\bibitem{russell08}
\BIBentryALTinterwordspacing
B.~C. Russell, A.~Torralba, K.~P. Murphy, and W.~T. Freeman, ``Labelme: A
  database and web-based tool for image annotation,'' \emph{International
  Journal of Computer Vision}, vol.~77, no.~1, pp. 157--173, May 2008.
  [Online]. Available: \url{https://doi.org/10.1007/s11263-007-0090-8}
\BIBentrySTDinterwordspacing

\bibitem{alex12}
\BIBentryALTinterwordspacing
A.~Krizhevsky, I.~Sutskever, and G.~E. Hinton, ``Imagenet classification with
  deep convolutional neural networks,'' in \emph{Advances in Neural Information
  Processing Systems 25}, F.~Pereira, C.~J.~C. Burges, L.~Bottou, and K.~Q.
  Weinberger, Eds.\hskip 1em plus 0.5em minus 0.4em\relax Curran Associates,
  Inc., 2012, pp. 1097--1105. [Online]. Available:
  \url{http://papers.nips.cc/paper/4824-imagenet-classification-with-deep-convolutional-neural-networks.pdf}
\BIBentrySTDinterwordspacing

\bibitem{Russakovsky2015}
\BIBentryALTinterwordspacing
O.~Russakovsky, J.~Deng, H.~Su, J.~Krause, S.~Satheesh, S.~Ma, Z.~Huang,
  A.~Karpathy, A.~Khosla, M.~Bernstein, A.~C. Berg, and L.~Fei-Fei, ``Imagenet
  large scale visual recognition challenge,'' \emph{Int. J. Comput. Vision},
  vol. 115, no.~3, pp. 211--252, Dec. 2015. [Online]. Available:
  \url{http://dx.doi.org/10.1007/s11263-015-0816-y}
\BIBentrySTDinterwordspacing

\bibitem{zeiler14}
M.~D. Zeiler and R.~Fergus, ``Visualizing and understanding convolutional
  networks,'' in \emph{Computer Vision -- ECCV 2014}, D.~Fleet, T.~Pajdla,
  B.~Schiele, and T.~Tuytelaars, Eds.\hskip 1em plus 0.5em minus 0.4em\relax
  Cham: Springer International Publishing, 2014, pp. 818--833.

\bibitem{simonyan14}
\BIBentryALTinterwordspacing
K.~Simonyan and A.~Zisserman, ``Very deep convolutional networks for
  large-scale image recognition,'' \emph{CoRR}, vol. abs/1409.1556, 2014.
  [Online]. Available: \url{http://arxiv.org/abs/1409.1556}
\BIBentrySTDinterwordspacing

\bibitem{szegedy15}
C.~Szegedy, W.~Liu, Y.~Jia, P.~Sermanet, S.~Reed, D.~Anguelov, D.~Erhan,
  V.~Vanhoucke, and A.~Rabinovich, ``Going deeper with convolutions,'' in
  \emph{The IEEE Conference on Computer Vision and Pattern Recognition (CVPR)},
  June 2015.

\bibitem{he16}
K.~He, X.~Zhang, S.~Ren, and J.~Sun, ``Deep residual learning for image
  recognition,'' in \emph{The IEEE Conference on Computer Vision and Pattern
  Recognition (CVPR)}, June 2016.

\bibitem{Huang16}
\BIBentryALTinterwordspacing
G.~Huang, Z.~Liu, and K.~Q. Weinberger, ``Densely connected convolutional
  networks,'' \emph{CoRR}, vol. abs/1608.06993, 2016. [Online]. Available:
  \url{http://arxiv.org/abs/1608.06993}
\BIBentrySTDinterwordspacing

\bibitem{capsnet17}
\BIBentryALTinterwordspacing
S.~Sabour, N.~Frosst, and G.~E. Hinton, ``Dynamic routing between capsules,''
  \emph{CoRR}, vol. abs/1710.09829, 2017. [Online]. Available:
  \url{http://arxiv.org/abs/1710.09829}
\BIBentrySTDinterwordspacing

\bibitem{Hu17}
\BIBentryALTinterwordspacing
J.~Hu, L.~Shen, and G.~Sun, ``Squeeze-and-excitation networks,'' \emph{CoRR},
  vol. abs/1709.01507, 2017. [Online]. Available:
  \url{http://arxiv.org/abs/1709.01507}
\BIBentrySTDinterwordspacing

\bibitem{buhmann2000}
M.~D. Buhmann, ``Radial basis functions,'' \emph{Acta Numerica}, vol.~9, p.
  1–38, 2000.

\bibitem{mnist10}
\BIBentryALTinterwordspacing
Y.~LeCun and C.~Cortes, ``{MNIST} handwritten digit database,'' 2010. [Online].
  Available: \url{http://yann.lecun.com/exdb/mnist/}
\BIBentrySTDinterwordspacing

\bibitem{Leon10}
L.~Bottou, ``Large-scale machine learning with stochastic gradient descent,''
  in \emph{Proceedings of COMPSTAT'2010}, Y.~Lechevallier and G.~Saporta,
  Eds.\hskip 1em plus 0.5em minus 0.4em\relax Heidelberg: Physica-Verlag HD,
  2010, pp. 177--186.

\bibitem{nair10}
\BIBentryALTinterwordspacing
V.~Nair and G.~E. Hinton, ``Rectified linear units improve restricted boltzmann
  machines,'' in \emph{Proceedings of the 27th International Conference on
  International Conference on Machine Learning}, ser. ICML'10.\hskip 1em plus
  0.5em minus 0.4em\relax USA: Omnipress, 2010, pp. 807--814. [Online].
  Available: \url{http://dl.acm.org/citation.cfm?id=3104322.3104425}
\BIBentrySTDinterwordspacing

\bibitem{ILSVRC15}
O.~Russakovsky, J.~Deng, H.~Su, J.~Krause, S.~Satheesh, S.~Ma, Z.~Huang,
  A.~Karpathy, A.~Khosla, M.~Bernstein, A.~C. Berg, and L.~Fei-Fei, ``{ImageNet
  Large Scale Visual Recognition Challenge},'' \emph{International Journal of
  Computer Vision (IJCV)}, vol. 115, no.~3, pp. 211--252, 2015.

\bibitem{hinton12}
\BIBentryALTinterwordspacing
G.~E. Hinton, N.~Srivastava, A.~Krizhevsky, I.~Sutskever, and R.~Salakhutdinov,
  ``Improving neural networks by preventing co-adaptation of feature
  detectors,'' \emph{CoRR}, vol. abs/1207.0580, 2012. [Online]. Available:
  \url{http://arxiv.org/abs/1207.0580}
\BIBentrySTDinterwordspacing

\bibitem{Jolliffe2011}
\BIBentryALTinterwordspacing
I.~Jolliffe, \emph{Principal Component Analysis}.\hskip 1em plus 0.5em minus
  0.4em\relax Berlin, Heidelberg: Springer Berlin Heidelberg, 2011, pp.
  1094--1096. [Online]. Available:
  \url{https://doi.org/10.1007/978-3-642-04898-2_455}
\BIBentrySTDinterwordspacing

\bibitem{zeiler11}
M.~D. Zeiler, G.~W. Taylor, and R.~Fergus, ``Adaptive deconvolutional networks
  for mid and high level feature learning,'' in \emph{2011 International
  Conference on Computer Vision}, Nov 2011, pp. 2018--2025.

\bibitem{caltech101}
L.~Fei-Fei, R.~Fergus, and P.~Perona, ``Learning generative visual models from
  few training examples: An incremental bayesian approach tested on 101 object
  categories,'' in \emph{2004 Conference on Computer Vision and Pattern
  Recognition Workshop}, June 2004, pp. 178--178.

\bibitem{caltech256}
\BIBentryALTinterwordspacing
G.~Griffin, A.~Holub, and P.~Perona, ``Caltech256 image dataset,'' 2006.
  [Online]. Available:
  \url{http://www.vision.caltech.edu/Image_Datasets/Caltech256/}
\BIBentrySTDinterwordspacing

\bibitem{pascal2012}
M.~Everingham, L.~Van~Gool, C.~K.~I. Williams, J.~Winn, and A.~Zisserman,
  ``Pascal visual object classes challenge 2012 (voc2012) complete dataset.''

\bibitem{Lin13}
\BIBentryALTinterwordspacing
M.~Lin, Q.~Chen, and S.~Yan, ``Network in network,'' \emph{CoRR}, vol.
  abs/1312.4400, 2013. [Online]. Available:
  \url{http://arxiv.org/abs/1312.4400}
\BIBentrySTDinterwordspacing

\bibitem{serre07}
T.~Serre, L.~Wolf, S.~Bileschi, M.~Riesenhuber, and T.~Poggio, ``Robust object
  recognition with cortex-like mechanisms,'' \emph{IEEE Transactions on Pattern
  Analysis and Machine Intelligence}, vol.~29, no.~3, pp. 411--426, March 2007.

\bibitem{dean12}
\BIBentryALTinterwordspacing
J.~Dean, G.~Corrado, R.~Monga, K.~Chen, M.~Devin, M.~Mao, M.~aurelio Ranzato,
  A.~Senior, P.~Tucker, K.~Yang, Q.~V. Le, and A.~Y. Ng, ``Large scale
  distributed deep networks,'' in \emph{Advances in Neural Information
  Processing Systems 25}, F.~Pereira, C.~J.~C. Burges, L.~Bottou, and K.~Q.
  Weinberger, Eds.\hskip 1em plus 0.5em minus 0.4em\relax Curran Associates,
  Inc., 2012, pp. 1223--1231. [Online]. Available:
  \url{http://papers.nips.cc/paper/4687-large-scale-distributed-deep-networks.pdf}
\BIBentrySTDinterwordspacing

\bibitem{CIFAR}
A.~Krizhevsky and G.~Hinton, ``Learning multiple layers of features from tiny
  images,'' Tech. Rep., 2009.

\bibitem{SVHN11}
\BIBentryALTinterwordspacing
Y.~Netzer, T.~Wang, A.~Coates, A.~Bissacco, B.~Wu, and A.~Y. Ng, ``Reading
  digits in natural images with unsupervised feature learning,'' in \emph{NIPS
  Workshop on Deep Learning and Unsupervised Feature Learning 2011}, 2011.
  [Online]. Available:
  \url{http://ufldl.stanford.edu/housenumbers/nips2011_housenumbers.pdf}
\BIBentrySTDinterwordspacing

\bibitem{sutskever13}
\BIBentryALTinterwordspacing
I.~Sutskever, J.~Martens, G.~Dahl, and G.~Hinton, ``On the importance of
  initialization and momentum in deep learning,'' in \emph{Proceedings of the
  30th International Conference on Machine Learning}, ser. Proceedings of
  Machine Learning Research, S.~Dasgupta and D.~McAllester, Eds., vol.~28,
  no.~3.\hskip 1em plus 0.5em minus 0.4em\relax Atlanta, Georgia, USA: PMLR,
  17--19 Jun 2013, pp. 1139--1147. [Online]. Available:
  \url{http://proceedings.mlr.press/v28/sutskever13.html}
\BIBentrySTDinterwordspacing

\bibitem{CIFAR10}
\BIBentryALTinterwordspacing
A.~Krizhevsky, V.~Nair, and G.~Hinton, ``Cifar-10 (canadian institute for
  advanced research).'' [Online]. Available:
  \url{http://www.cs.toronto.edu/~kriz/cifar.html}
\BIBentrySTDinterwordspacing

\bibitem{CIFAR100}
\BIBentryALTinterwordspacing
A.~Krizhevsky, V.~Nair, and G.~E. Hinton, ``Cifar-100 (canadian institute for
  advanced research).'' [Online]. Available:
  \url{http://www.cs.toronto.edu/~kriz/cifar.html}
\BIBentrySTDinterwordspacing

\bibitem{Xie16}
\BIBentryALTinterwordspacing
S.~Xie, R.~B. Girshick, P.~Doll{\'{a}}r, Z.~Tu, and K.~He, ``Aggregated
  residual transformations for deep neural networks,'' \emph{CoRR}, vol.
  abs/1611.05431, 2016. [Online]. Available:
  \url{http://arxiv.org/abs/1611.05431}
\BIBentrySTDinterwordspacing

\bibitem{SzegedyIV16}
\BIBentryALTinterwordspacing
C.~Szegedy, S.~Ioffe, and V.~Vanhoucke, ``Inception-v4, inception-resnet and
  the impact of residual connections on learning,'' \emph{CoRR}, vol.
  abs/1602.07261, 2016. [Online]. Available:
  \url{http://arxiv.org/abs/1602.07261}
\BIBentrySTDinterwordspacing

\bibitem{Howard17}
\BIBentryALTinterwordspacing
A.~G. Howard, M.~Zhu, B.~Chen, D.~Kalenichenko, W.~Wang, T.~Weyand,
  M.~Andreetto, and H.~Adam, ``Mobilenets: Efficient convolutional neural
  networks for mobile vision applications,'' \emph{CoRR}, vol. abs/1704.04861,
  2017. [Online]. Available: \url{http://arxiv.org/abs/1704.04861}
\BIBentrySTDinterwordspacing

\bibitem{Zhang17}
\BIBentryALTinterwordspacing
X.~Zhang, X.~Zhou, M.~Lin, and J.~Sun, ``Shufflenet: An extremely efficient
  convolutional neural network for mobile devices,'' \emph{CoRR}, vol.
  abs/1707.01083, 2017. [Online]. Available:
  \url{http://arxiv.org/abs/1707.01083}
\BIBentrySTDinterwordspacing

\end{thebibliography}

\end{document}